\documentclass[twocolumn,
]{ceurart}
\usepackage[utf8]{inputenc}
\usepackage[frozencache=true,cachedir=minted-cache]{minted} 
\setminted{breaklines=true}
\begin{document}

\copyrightyear{2023}
\copyrightclause{Copyright for this paper by its authors.
  Use permitted under Creative Commons License Attribution 4.0
  International (CC BY 4.0).}

\conference{SEPLN 2023: 39\textsuperscript{th} International Conference of the Spanish Society for Natural Language Processing}

\title{HiTZ@Antidote: Argumentation-driven Explainable Artificial Intelligence for Digital Medicine}

\author{Rodrigo Agerri}[%
%orcid=0000-0002-7303-7598,
email=rodrigo.agerri@ehu.eus
]
\address{HiTZ Center - Ixa, University of the Basque Country UPV/EHU}

\author{I\~nigo Alonso}

\author{Aitziber Atutxa}[%
%orcid=0000-0000-0000-0000
]

\author{Ander Berrondo}[%
%orcid=0000-0000-0000-0000
]

\author{Ainara Estarrona}

\author{Iker Garcia-Ferrero}

\author{Iakes Goenaga}[
%orcid=0000-0000-0000-0000
]
\author{Koldo Gojenola}[
%orcid=0000-0000-0000-0000
]

\author{Maite Oronoz}[
%orcid=0000-0000-0000-0000
]
\author{Igor Perez-Tejedor}

\author{German Rigau}[
%orcid=0000-0003-1119-0930
]

\author{Anar Yeginbergenova}

\begin{abstract}
Providing high quality explanations for AI predictions based on machine
learning is a challenging and complex task. To work well it requires, among
other factors: selecting a proper level of generality/specificity of the
explanation; considering assumptions about the familiarity of the explanation
beneficiary with the AI task under consideration; referring to specific
elements that have contributed to the decision; making use of additional
knowledge (e.g.  expert evidence) which might not be part of the prediction
process; and providing evidence supporting negative hypothesis. Finally, the system
needs to formulate the explanation in a clearly interpretable, and possibly
convincing, way.  Given these considerations, ANTIDOTE fosters an integrated
vision of explainable AI, where low-level characteristics of the deep learning
process are combined with higher level schemes proper of the human
argumentation capacity. ANTIDOTE will exploit cross-disciplinary competences in
deep learning and argumentation to support a broader and
innovative view of explainable AI, where the need for high-quality
explanations for clinical cases deliberation is critical. As a first result of
the project, we publish the \emph{Antidote CasiMedicos} dataset to facilitate
research on explainable AI in general, and argumentation in the medical domain in particular. 
 \end{abstract}


\maketitle

\section{Introduction}

ANTIDOTE\footnote{\url{https://univ-cotedazur.eu/antidote}} is a European CHIST-ERA project where each partner is funded by their national Science Agencies. As the Spanish partner in the Consortium is the HiTZ Center - Ixa, from the University of the Basque Country UPV/EHU, the project was funded by the \emph{Proyectos de Colaboración Internacional} (PCI 2020) program of the Spanish Ministry of Science and Innovation. The other European partners are the following: Université Côte d’Azur (UCA) from France and coordinators of the international consortium, Fondazione Bruno Kessler (FBK) from Italy, KU Leuven/Computer Science, in Belgium and Universidade Nova de Lisboa (NOVA) in Portugal. 

The aim of ANTIDOTE is to exploit cross-disciplinary competences in three
areas, namely, deep learning, argumentation and interactivity, to support a
broader and innovative view of explainable AI. 

Providing high quality explanations for AI predictions based on machine
learning is a challenging and complex task. To work well it requires, among
other aspects: (i) selecting a proper level of generality/specificity of the
explanation, (ii) considering assumptions about the familiarity of the
explanation beneficiary with the AI task under consideration, (iii) referring
to specific elements that have contributed to the decision, (iv) making use of
additional knowledge (e.g. metadata) which might not be part of the prediction
process, (v) selecting appropriate examples and, (vi) providing evidence
supporting negative hypotheses. Finally, the system needs to formulate the
explanation in a clearly interpretable, and possibly convincing, way.

Taking into account these considerations, ANTIDOTE fosters an integrated vision
of Explainable AI (XAI), where the low-level characteristics of the deep learning
process are combined with higher level schemes proper of human
argumentation. Following this, the ANTIDOTE integrated vision is supported by
three considerations. First, in neural architectures the correlation between
internal states of the network (e.g., weights assumed by single nodes) and the
justification of the network classification outcome is not well studied.
Second, high quality explanations are crucially based on argumentation
mechanisms (e.g., provide supporting examples and rejected alternatives).
Finally, in real settings, providing explanations is inherently an interactive
process involving the system and the user.

Thus, ANTIDOTE will exploit cross-disciplinary
competences in three areas, namely, deep learning, argumentation and
interactivity, to support a broader and innovative view of explainable AI.
There are several research challenges that ANTIDOTE will address to advance the
state-of-the-art in explainable AI.

The first challenge is to take advantage of the huge body of past research on
argumentation to complement state-of-the-art approaches on explainability. In
addition, the recent resurgence of AI highlights the idea that low-level system
behavior not only needs to be interpretable (e.g., showing those elements that
most contributed to the system decision), but that also needs to be joined by
high level human argumentation schemes. The second challenge is to
automatically learn explanatory argumentation schemas in Natural Language (NL)
and to effectively combine evidence-based decision making with high level
explanations. The third challenge for ANTIDOTE is that a task-specific
prediction model and a general argumentation model need to be combined to
produce explanatory argumentations.

While neural networks for medical diagnosis have become exceedingly accurate
in many areas, their ability to explain how they achieve their outcome remains
problematic. Herein lies the main novelty of the ANTIDOTE project: it 
focuses on elaborating argumentative explanations to diagnosis predictions in
order to assist student clinicians to learn making informed decisions. 

The explanatory argumentative scenario envisaged by ANTIDOTE will involve a
student clinician, who will need to hypothesize about the clinical case of a patient
and will have to provide argumentative explanations about them. 
The focus of the experimental setting is set on the capacity of the
ANTIDOTE Explanatory AI system to provide correct predictions and consistent
arguments, without forgetting also the linguistic quality of the dialogues
(e.g., naturalness of the utterances, etc.).

\begin{figure}
  \centering
  \includegraphics[width=\linewidth]{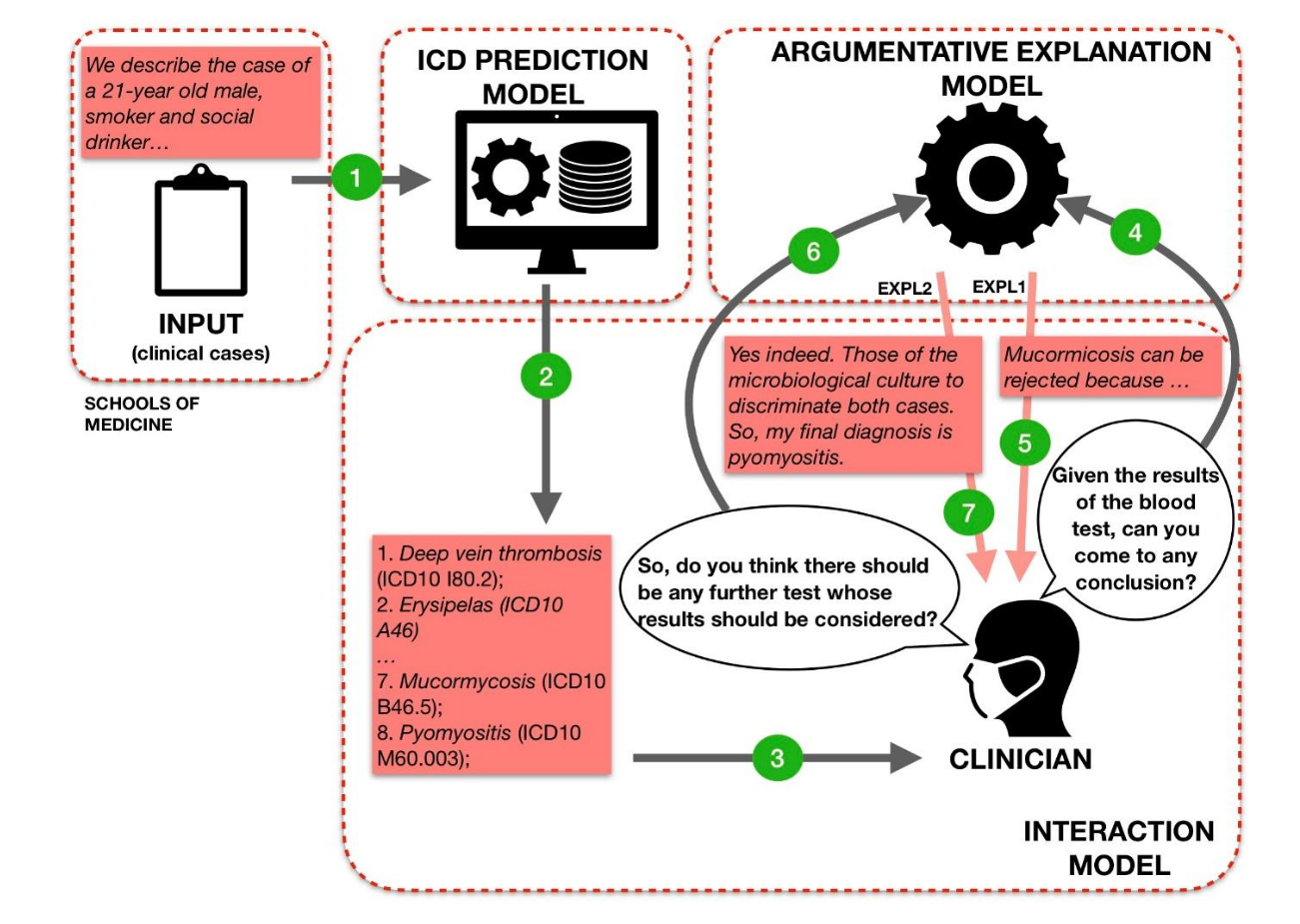}
  \caption{ANTIDOTE use-case scenario.}
  \label{fig:scenario}
\end{figure}

In our scenario depicted in Figure \ref{fig:scenario}, the clinician queries the ANTIDOTE XAI for
explanations (arguments) on its diagnosis of the clinical case. The ANTIDOTE
XAI provides hypotheses (differential diagnosis) about the clinical case, as
well as arguments to support its prediction and arguments discarding
alternative predictions. The student clinician has the possibility to take the
initiative to ask additional questions and clarifications. The goal of the
explanatory argumentation in a differential diagnosis is to validate the correctness
of the diagnosis and the ANTIDOTE XAI capacity to argue in favour of the
correct hypothesis and to counter-argue against alternative hypotheses. 

\section{Related Work}\label{sec:related-work}

In this section we review the most relevant previous work focusing on
argumentation and explainable AI for the medical domain.

\subsection{Argumentation mining and generation}

Argumentation mining is a research area that moves between natural language
processing, argumentation theory and information retrieval. The aim of
argumentation mining is to automatically detect the argumentation of a document
and its structure. This implies the detection of all the arguments involved in
the argumentation process, their individual or local structure (rhetorical or
argumentative relationships between their propositions), and the interactions
between them, namely, the global argumentation structure.

Argumentation mining in Natural Language Processing has been applied to various
domains such as persuasive essays, legal documents, political debates and
social media data \citep{dusmanu2017argument}. For instance, Stab and Gurevych
\cite{stab2017parsing} built an annotated dataset of persuasive essays with
corresponding argument components and relations. Using this corpus, Eger et al.
\cite{eger2017neural} developed an end-to-end neural method for argument
structure identification. Furthermore, Nguyen and Litman
\cite{nguyen2018argument} also applied an end-to-end method to parse argument
structure and used the argument structure features to improve automated
persuasive essay scoring. Other approaches studied context-dependent claim
detection by collecting annotations for Wikipedia articles
\cite{levy2014context}. Using this corpus, the task of automatically
identifying the corresponding pieces of evidence given a claim has also been
investigated \cite{rinott2015show}.

Argumentation generation remains a research area in which there
is still a long way to go. Recent work has made progress towards this goal
through the automated generation of argumentative text
\citep{bar2020arguments,hua2018neural,sato2015end,alshomary2021argument}.
Thus, Alshomary et al. \cite{alshomary2020target}
proposed a Bayesian argument generation system to generate arguments given
the corresponding argumentation strategies. Sato et al. \cite{sato2015end} presented a
sentence-retrieval-based end-to-end argument generation system that can
participate in English debating games.

There have also been some works exploring counter-argument generation to
select the main talking points to generate a counter-argument
\cite{hua2019argument}. In this line of research, Hidey and McKeown
\cite{hidey2019fixed} proposed a neural model that edited
the original claim semantically to produce a claim with an opposing stance.
They also incorporated external knowledge into the encoder-decoder architecture
showing that their model generated arguments that were more likely to be on topic. 

Finally, an autonomous debating system (Project
Debater) able to engage in competitive debates with humans was developed. The
system consisted of a pipeline of four main modules: argument mining, an argument knowledge base, argument rebuttal, and debate construction \cite{slonim2021autonomous}.

\subsection{Explainable AI}

\begin{figure}
  \centering
  \includegraphics[width=\linewidth]{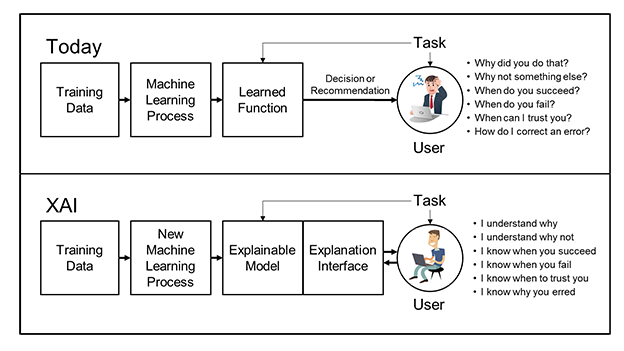}
  \caption{Explainable AI with Human in the Loop.}
  \label{fig:xai-darpa}
\end{figure}

Explainable artificial intelligence (XAI) aims to address the needs of users 
wanting to understand how a program's artificial intelligence
works and how to evaluate the results obtained. Otherwise, there is no basis
for real confidence in the work of the AI system, as illustrated by Figure
\ref{fig:xai-darpa}\footnote{Source from DARPA XAI
  program: \url{https://www.darpa.mil/program/explainable-artificial-intelligence}}
The transparency offered by
explainable AI is therefore essential for the acceptance of artificial
intelligence.

There has been a surge of interest in explainable artificial intelligence (XAI)
in recent years. This has produced a myriad of algorithmic and
mathematical methods to explain the inner workings of machine learning models
\citep{biran2017explanation}. However, despite their
mathematical rigor, these works suffer from a lack of usability and practical
interpretability for real users. Although the concepts of
interpretability and explainability are hard to rigorously define, multiple
attempts have been made towards that goal
\citep{lipton2018mythos,doshi2017towards}.

Adadi and Berranda \cite{adadi2018peeking} presented an extensive literature review, collecting
and analyzing 381 different scientific papers between 2004 and 2018. They
arranged all of the scientific work in the field of explainable AI along four
main axes and stressed the need for more formalism to be introduced in the
field of XAI and for more interaction between humans and machines.

In a more recent study \cite{arrieta2020explainable} introduced a different
type of arrangement that initially distinguishes transparent and post-hoc
methods and subsequently created sub-categories.

Taking into account argumentation principles, ANTIDOTE will explain machine
decisions based on four modes of explanations to be auditable by humans: (i) analytic
statements in NL that describe the elements and context that support a choice, (ii) visualizations
that highlight portions of the raw data that support a choice, (iii) cases that invoke specific
examples, and (iv) rejections of alternative choices that argue against less preferred answers
based on analytics, cases, and data.

\section{Methology and Work Plan}

The main scientific challenge for the project is the combination of three
models depicted in Figure \ref{fig:scenario}: (1) The Prediction Model has to
predict appropriate International Classification of Diseases (ICD) codes given a clinical case; 
(2) the Argumentative
Model selects proper arguments (i.e., entity and relations) to support or 
attack a given topic. It may use both information included in the clinical
cases used by the prediction model and additional sources of knowledge; (3) the
Interaction Model provides argumentative explanations about a certain
prediction. An integrated approach is proposed to both predict the outcome of a
clinical course of action and justify a medical diagnosis by a language model.
A starting point will be using current large language models
\cite{xue-etal-2021-mt5} to generate appropriate explanations guided by the
activated view on a textual snippet that contributed to the decision, namely,
the argument for the decision.

\subsection{Work Plan}

The Work Plan is structured in six Work Packages of which three are focused on
the scientific contributions of the project.

\begin{itemize}
\item[\textbf{WP2}:] Methodology and Design (Leader: FBK). Participants: UCA, UPV/EHU, KU, NOVA.
The purpose of WP2 is to define, adapt and integrate the modules, resources, data structures,
data formats and module APIs of the ANTIDOTE architecture. This includes designing the
experiments, datasets, standard protocols, information flow and main architecture of
ANTIDOTE. 

\item[\textbf{WP3}:] Machine Learning (ML) for predicting clinical outcomes (Leader: KU).  Participants: UPV/EHU, FBK, UCA, NOVA.  WP3 targets (1) the development of a multitask learning model to jointly predict and justify a
medical diagnosis by a deep learning model; (2) surfacing and making explicit the underlying
aspects (identification of the most relevant/informative terms, identification of relations
among terms) driving neural network decisions during the diagnosis prediction process; (3)
retrieve external information to support the explanation.

\item[\textbf{WP4}:] Explanatory arguments in natural language (Leader:
UCA). Participants: UPV/EHU, FBK, UCA, NOVA.
WP4 relies on the textual arguments that form the basis for the decisions
generated in WP3. WP4 targets (1) the definition and analysis of explanatory argumentative
patterns to be used to construct natural language explanatory arguments of
predictions; (2) the creation of a resource of annotated natural language explanatory
arguments; (3) the development of explanatory arguments
in natural language by mining and collecting them from trusted textual resources in the medical
domain.

\item[\textbf{WP5}:] Evaluation (use cases in healthcare) (Leader: UPV/EHU).
Participants: KU, UCA, FBK, NOVA.
WP5 aims to (1) evaluate the effectiveness and quality of the prediction and the plausible
alternatives (2) the quality of the generated explanatory arguments regarding the supporting
evidence found in the clinical case in favor of the prediction and the positive or negative
evidence found to discard other plausible alternatives, (3) the intrinsic
quality of the generated arguments.
\end{itemize}

\subsection{Evaluation}

The generation of arguments will be quantitatively evaluated by computing
metrics used in text generation to measure their overlap with ground truth
arguments \cite{zhangbertscore}. Moreover, the argumentative model will be
evaluated following the criteria of coherence, simplicity, and generality
\cite{miller2019explanation}: explanations with structural simplicity,
coherence, or minimality are preferred. With respect to argument mining,
standard metrics such as F1 and accuracy will be used.

Generation of explanatory arguments will be also qualitative evaluated by
medical students. Given the objectives and context of the project, ANTIDOTE
will be based on previous work by Johnson \cite{johnson2012manifest},
whereby the arguments will be evaluated for their (informal) inferential
structure in terms of acceptability, relevance, and sufficiency of reasons
provided, as well as their answerability to human agents' doubts and
objections.

\section{Ongoing Work}

There are a number of tasks currently being undertaken within the project. In
this section we provide details of the most central ones with respect to the
objectives and motivation provided in the introduction.

\subsection{ANTIDOTE Datasets}

In order to carry out the tasks related to the main use-case presented in
Figure \ref{fig:scenario}, we need to identify, collect and annotate the most
suitable corpus with which to train different models. In this regard, we have
identified two possible data sources that will help us meet our objectives and
that will constitute an important contribution of the ANTIDOTE project: 
SAEI and CasiMedicos. 

\paragraph{The SAEI Corpus} is a collection of differential diagnosis in
Spanish collected by \textit{La Sociedad Andaluza de Enfermedades Infecciosas} (The Andalusian
Society of Infectious Diseases)\footnote{\url{https://www.saei.org}}. This society is a non-profit association formed almost entirely by physicians specializing in Internal Medicine with
special dedication to the management of infectious diseases, whose general
purpose is the promotion and development of this medical discipline (training,
care and research). We have selected the books that
are of interest to us in order to carry out our objectives, namely, those that include clinical cases of infectious diseases for residents that are available, for the years 2011, 2015, 2016, 2017 and 2020. Among all these books we have extracted cleaned and pre-processed a total of 244 clinical cases with differential diagnosis.
%\footnote{https://github.com/ixa-ehu/antidote-saei}

\paragraph{CasiMedicos} is a community and collaborative medical project run by
volunteer medical doctors\footnote{\url{https://www.casimedicos.com/}}. 
Among all the information created and made publicly available by this
collaborative project, we have identified as an adequate data source the MIR
exams commented by voluntary medical doctors with the aim of providing
answers and explanations\footnote{\url{https://www.casimedicos.com/mir-2-0/}} to the MIR exams annually published by the Spanish Ministry of Health. In this data source we have
extracted and pre-processed 622 commented questions from the MIR
exams held between the years 2005, 2014, 2016, 2018, 2019, 2020, 2021 and 2022. The cleaned corpus, named the \emph{Antidote Casimedicos} dataset, is publicly available to encourage research on explainable AI in the medical domain in general, and argumentation in particular\footnote{\url{https://github.com/ixa-ehu/antidote-casimedicos}}.

Unlike popular Question Anwering (QA) datasets for English based on medical exams
\cite{singhal2022large}, both SAEI and CasiMedicos include not only the
explanations for the corrent answer (diagnosis or treatment), but also explanatory arguments written by
medical doctors explaining why the rest of the possible answers are incorrect.

After pre-processing, these datasets have been translated from Spanish to
English with the objective of starting various annotation tasks at various
levels of complexity: (i) linking the explanatory sequences with respect to
each possible answer; (ii) labeling of hierarchical argumentative structures;
(iii) discourse markers. The resulting corpus will be the first corpus
(multilingual or otherwise) with this type of annotations for the medical
domain. Whenever ready, the corpus will be distributed under a free license to
promote further research and to ensure reproducibility or results. 

\subsection{Question Answering in the Medical Domain}

While there are several QA datasets for English based on medical exams
\cite{singhal2022large}, none of the previously published works contain two
features which are unique of both SAEI and CasiMedicos: (i) the presence of
explanations for both correct and incorrect answers; (ii) an argumentative
structure arguing and counter-arguing about the possible answers. These
features make it possible to define new Question Answering tasks, both from an
extractive and generative point of view. In extractive QA, the objective would
consist of identifying, in a given context, the explanation to the correct answer.
In terms of generative QA, it will also allow us to leverage large
language models \cite{xue-etal-2021-mt5} to learn generating the explanatory
arguments with respect to both correct and incorrect possible answers.

\subsection{Crosslingual Knowledge Transfer}

The only corpus annotated with argumentative structure currently available for
the medical domain is the AbstRCT dataset, which consists of English clinical
trials \cite{mayer2021enhancing}. In order to investigate the different
strategies of transferring knowledge from English to other languages,
especially those applying model- and data-transfer techniques previously discussed for
other application domains \cite{garcia-ferrero-etal-2022-model}, ongoing work is focused on adapting such knowledge transfer techniques for argumentation in the medical domain. As a result,
we are undertaking novel experimental work on argument mining in Spanish
for the medical domain \cite{yeginbergenova2023crosslingual}. This also involves the 
generation of the first Spanish dataset annotated with argumentative structures for the medical domain. Finally, the
plan is to apply the developed technique to other languages of interest for
the ANTIDOTE project (French and Italian).

In this line of research, and taking as starting point the ongoing work
mentioned in the previous section, we plan to investigate also crosslingual and
multilingual approaches to Question Answering techniques in the medical domain.

\section{Concluding Remarks}

In this paper we provide a description of the ANTIDOTE project, mostly focusing on
identifying and generating high-quality argumentative explanations for AI
predictions in the medical domain. So far, ongoing work has been focused on
dataset collection and annotation and novel experimental work on Question
Answering and Crosslingual Argument Mining. This work has leveraged
multilingual encoder and decoder large language models
\cite{Devlin19,xue-etal-2021-mt5} for both extractive and generative
experimentation.

Still, providing high-quality explanations for AI predictions based on machine
learning is a challenging and complex task \cite{singhal2022large}. To work
well, it requires, among other factors, making use of additional knowledge (e.g.
medical evidence) which might not be part of the prediction process, and
providing evidence supporting negative hypotheses. With these issues in mind,
ANTIDOTE aims to address the challenge of providing an integrated vision of
explainable AI, where low-level characteristics of the deep learning process
are combined with higher level schemes proper of the human argumentation
capacity. In order to do so, ANTIDOTE will be focused on a number of deep
learning tasks for the medical domain, where the need for high quality
explanations for clinical cases deliberation is critical.

%\begin{table*}
%  \caption{Frequency of Special Characters}
%  \label{tab:freq}
%  \begin{tabular}{ccl}
%    \toprule
%    Non-English or Math&Frequency&Comments\\
%    \midrule
%    \O & 1 in 1,000& For Swedish names\\
%    $\pi$ & 1 in 5& Common in math\\
%    \$ & 4 in 5 & Used in business\\
%    $\Psi^2_1$ & 1 in 40,000& Unexplained usage\\
%  \bottomrule
%\end{tabular}
%\end{table*}

%\begin{table}
%  \caption{Some Typical Commands}
%  \label{tab:commands}
%  \begin{tabular}{ccl}
%    \toprule
%    Command &A Number & Comments\\
%    \midrule
%    \texttt{{\char'134}author} & 100& Author \\
%    \texttt{{\char'134}table}& 300 & For tables\\
%    \texttt{{\char'134}table*}& 400& For wider tables\\
%    \bottomrule
%  \end{tabular}
% \end{table}

\begin{acknowledgments}
We thank the CasiMedicos Proyecto MIR 2.0 for their permission to share their data for research purposes. ANTIDOTE (PCI2020-120717-2) is a project funded by MCIN/AEI/10.13039/501100011033 and by European Union NextGenerationEU/PRTR. Rodrigo Agerri currently holds the RYC-2017-23647 fellowship
(MCIN/AEI/10.13039/501100011033 and by ESF Investing in your future). Iker García-Ferrero is supported by a doctoral grant from the Basque Government (PRE\_2021\_2\_0219) and Anar Yeginbergenova acknowledges the PhD contract from the UPV/EHU (PIF 22/159).

\end{acknowledgments}

%% Define the bibliography file to be used
\bibliography{references}

\begin{thebibliography}{28}
\expandafter\ifx\csname natexlab\endcsname\relax\def\natexlab#1{#1}\fi
\providecommand{\url}[1]{\texttt{#1}}
\providecommand{\href}[2]{#2}
\providecommand{\path}[1]{#1}
\providecommand{\DOIprefix}{doi:}
\providecommand{\ArXivprefix}{arXiv:}
\providecommand{\URLprefix}{URL: }
\providecommand{\Pubmedprefix}{pmid:}
\providecommand{\doi}[1]{\href{http://dx.doi.org/#1}{\path{#1}}}
\providecommand{\Pubmed}[1]{\href{pmid:#1}{\path{#1}}}
\providecommand{\bibinfo}[2]{#2}
\ifx\xfnm\relax \def\xfnm[#1]{\unskip,\space#1}\fi
%Type = Inproceedings
\bibitem[{Dusmanu et~al.(2017)Dusmanu, Cabrio, and
  Villata}]{dusmanu2017argument}
\bibinfo{author}{M.~Dusmanu}, \bibinfo{author}{E.~Cabrio},
  \bibinfo{author}{S.~Villata},
\newblock \bibinfo{title}{Argument mining on twitter: Arguments, facts and
  sources},
\newblock in: \bibinfo{booktitle}{Proceedings of the 2017 Conference on
  Empirical Methods in Natural Language Processing}, \bibinfo{year}{2017}, pp.
  \bibinfo{pages}{2317--2322}.
%Type = Article
\bibitem[{Stab and Gurevych(2017)}]{stab2017parsing}
\bibinfo{author}{C.~Stab}, \bibinfo{author}{I.~Gurevych},
\newblock \bibinfo{title}{Parsing argumentation structures in persuasive
  essays},
\newblock \bibinfo{journal}{Computational Linguistics} \bibinfo{volume}{43}
  (\bibinfo{year}{2017}) \bibinfo{pages}{619--659}.
%Type = Article
\bibitem[{Eger et~al.(2017)Eger, Daxenberger, and Gurevych}]{eger2017neural}
\bibinfo{author}{S.~Eger}, \bibinfo{author}{J.~Daxenberger},
  \bibinfo{author}{I.~Gurevych},
\newblock \bibinfo{title}{Neural end-to-end learning for computational
  argumentation mining},
\newblock \bibinfo{journal}{arXiv preprint arXiv:1704.06104}
  (\bibinfo{year}{2017}).
%Type = Inproceedings
\bibitem[{Nguyen and Litman(2018)}]{nguyen2018argument}
\bibinfo{author}{H.~Nguyen}, \bibinfo{author}{D.~Litman},
\newblock \bibinfo{title}{Argument mining for improving the automated scoring
  of persuasive essays},
\newblock in: \bibinfo{booktitle}{Proceedings of the AAAI Conference on
  Artificial Intelligence}, volume~\bibinfo{volume}{32}, \bibinfo{year}{2018}.
%Type = Inproceedings
\bibitem[{Levy et~al.(2014)Levy, Bilu, Hershcovich, Aharoni, and
  Slonim}]{levy2014context}
\bibinfo{author}{R.~Levy}, \bibinfo{author}{Y.~Bilu},
  \bibinfo{author}{D.~Hershcovich}, \bibinfo{author}{E.~Aharoni},
  \bibinfo{author}{N.~Slonim},
\newblock \bibinfo{title}{Context dependent claim detection},
\newblock in: \bibinfo{booktitle}{Proceedings of COLING 2014, the 25th
  International Conference on Computational Linguistics: Technical Papers},
  \bibinfo{year}{2014}, pp. \bibinfo{pages}{1489--1500}.
%Type = Inproceedings
\bibitem[{Rinott et~al.(2015)Rinott, Dankin, Alzate, Khapra, Aharoni, and
  Slonim}]{rinott2015show}
\bibinfo{author}{R.~Rinott}, \bibinfo{author}{L.~Dankin},
  \bibinfo{author}{C.~Alzate}, \bibinfo{author}{M.~M. Khapra},
  \bibinfo{author}{E.~Aharoni}, \bibinfo{author}{N.~Slonim},
\newblock \bibinfo{title}{Show me your evidence-an automatic method for context
  dependent evidence detection},
\newblock in: \bibinfo{booktitle}{Proceedings of the 2015 conference on
  empirical methods in natural language processing}, \bibinfo{year}{2015}, pp.
  \bibinfo{pages}{440--450}.
%Type = Article
\bibitem[{Bar-Haim et~al.(2020)Bar-Haim, Eden, Friedman, Kantor, Lahav, and
  Slonim}]{bar2020arguments}
\bibinfo{author}{R.~Bar-Haim}, \bibinfo{author}{L.~Eden},
  \bibinfo{author}{R.~Friedman}, \bibinfo{author}{Y.~Kantor},
  \bibinfo{author}{D.~Lahav}, \bibinfo{author}{N.~Slonim},
\newblock \bibinfo{title}{From arguments to key points: Towards automatic
  argument summarization},
\newblock \bibinfo{journal}{arXiv preprint arXiv:2005.01619}
  (\bibinfo{year}{2020}).
%Type = Article
\bibitem[{Hua and Wang(2018)}]{hua2018neural}
\bibinfo{author}{X.~Hua}, \bibinfo{author}{L.~Wang},
\newblock \bibinfo{title}{Neural argument generation augmented with externally
  retrieved evidence},
\newblock \bibinfo{journal}{arXiv preprint arXiv:1805.10254}
  (\bibinfo{year}{2018}).
%Type = Inproceedings
\bibitem[{Sato et~al.(2015)Sato, Yanai, Miyoshi, Yanase, Iwayama, Sun, and
  Niwa}]{sato2015end}
\bibinfo{author}{M.~Sato}, \bibinfo{author}{K.~Yanai},
  \bibinfo{author}{T.~Miyoshi}, \bibinfo{author}{T.~Yanase},
  \bibinfo{author}{M.~Iwayama}, \bibinfo{author}{Q.~Sun},
  \bibinfo{author}{Y.~Niwa},
\newblock \bibinfo{title}{End-to-end argument generation system in debating},
\newblock in: \bibinfo{booktitle}{Proceedings of ACL-IJCNLP 2015 System
  Demonstrations}, \bibinfo{year}{2015}, pp. \bibinfo{pages}{109--114}.
%Type = Article
\bibitem[{Alshomary et~al.(2021)Alshomary, Syed, Dhar, Potthast, and
  Wachsmuth}]{alshomary2021argument}
\bibinfo{author}{M.~Alshomary}, \bibinfo{author}{S.~Syed},
  \bibinfo{author}{A.~Dhar}, \bibinfo{author}{M.~Potthast},
  \bibinfo{author}{H.~Wachsmuth},
\newblock \bibinfo{title}{{Argument Undermining: Counter-Argument Generation by
  Attacking Weak Premises}},
\newblock \bibinfo{journal}{arXiv preprint arXiv:2105.11752}
  (\bibinfo{year}{2021}).
%Type = Inproceedings
\bibitem[{Alshomary et~al.(2020)Alshomary, Syed, Potthast, and
  Wachsmuth}]{alshomary2020target}
\bibinfo{author}{M.~Alshomary}, \bibinfo{author}{S.~Syed},
  \bibinfo{author}{M.~Potthast}, \bibinfo{author}{H.~Wachsmuth},
\newblock \bibinfo{title}{Target inference in argument conclusion generation},
\newblock in: \bibinfo{booktitle}{Proceedings of the 58th Annual Meeting of the
  Association for Computational Linguistics}, \bibinfo{year}{2020}, pp.
  \bibinfo{pages}{4334--4345}.
%Type = Article
\bibitem[{Hua et~al.(2019)Hua, Hu, and Wang}]{hua2019argument}
\bibinfo{author}{X.~Hua}, \bibinfo{author}{Z.~Hu}, \bibinfo{author}{L.~Wang},
\newblock \bibinfo{title}{Argument generation with retrieval, planning, and
  realization},
\newblock \bibinfo{journal}{arXiv preprint arXiv:1906.03717}
  (\bibinfo{year}{2019}).
%Type = Inproceedings
\bibitem[{Hidey and McKeown(2019)}]{hidey2019fixed}
\bibinfo{author}{C.~Hidey}, \bibinfo{author}{K.~McKeown},
\newblock \bibinfo{title}{{Fixed that for you: Generating contrastive claims
  with semantic edits}},
\newblock in: \bibinfo{booktitle}{Proceedings of the 2019 Conference of the
  North American Chapter of the Association for Computational Linguistics:
  Human Language Technologies, Volume 1 (Long and Short Papers)},
  \bibinfo{year}{2019}, pp. \bibinfo{pages}{1756--1767}.
%Type = Article
\bibitem[{Slonim et~al.(2021)Slonim, Bilu, Alzate, Bar-Haim, Bogin, Bonin,
  Choshen, Cohen-Karlik, Dankin, Edelstein et~al.}]{slonim2021autonomous}
\bibinfo{author}{N.~Slonim}, \bibinfo{author}{Y.~Bilu},
  \bibinfo{author}{C.~Alzate}, \bibinfo{author}{R.~Bar-Haim},
  \bibinfo{author}{B.~Bogin}, \bibinfo{author}{F.~Bonin},
  \bibinfo{author}{L.~Choshen}, \bibinfo{author}{E.~Cohen-Karlik},
  \bibinfo{author}{L.~Dankin}, \bibinfo{author}{L.~Edelstein}, et~al.,
\newblock \bibinfo{title}{An autonomous debating system},
\newblock \bibinfo{journal}{Nature} \bibinfo{volume}{591}
  (\bibinfo{year}{2021}) \bibinfo{pages}{379--384}.
%Type = Inproceedings
\bibitem[{Biran and Cotton(2017)}]{biran2017explanation}
\bibinfo{author}{O.~Biran}, \bibinfo{author}{C.~Cotton},
\newblock \bibinfo{title}{Explanation and justification in machine learning: A
  survey},
\newblock in: \bibinfo{booktitle}{IJCAI-17 workshop on explainable AI (XAI)},
  volume~\bibinfo{volume}{8}, \bibinfo{year}{2017}, pp. \bibinfo{pages}{8--13}.
%Type = Article
\bibitem[{Lipton(2018)}]{lipton2018mythos}
\bibinfo{author}{Z.~C. Lipton},
\newblock \bibinfo{title}{{The Mythos of Model Interpretability: In machine
  learning, the concept of interpretability is both important and slippery.}},
\newblock \bibinfo{journal}{Queue} \bibinfo{volume}{16} (\bibinfo{year}{2018})
  \bibinfo{pages}{31--57}.
%Type = Article
\bibitem[{Doshi-Velez and Kim(2017)}]{doshi2017towards}
\bibinfo{author}{F.~Doshi-Velez}, \bibinfo{author}{B.~Kim},
\newblock \bibinfo{title}{Towards a rigorous science of interpretable machine
  learning},
\newblock \bibinfo{journal}{arXiv preprint arXiv:1702.08608}
  (\bibinfo{year}{2017}).
%Type = Article
\bibitem[{Adadi and Berrada(2018)}]{adadi2018peeking}
\bibinfo{author}{A.~Adadi}, \bibinfo{author}{M.~Berrada},
\newblock \bibinfo{title}{{Peeking inside the black-box: a survey on
  explainable artificial intelligence (XAI)}},
\newblock \bibinfo{journal}{IEEE access} \bibinfo{volume}{6}
  (\bibinfo{year}{2018}) \bibinfo{pages}{52138--52160}.
%Type = Article
\bibitem[{Arrieta et~al.(2020)Arrieta, D{\'\i}az-Rodr{\'\i}guez, Del~Ser,
  Bennetot, Tabik, Barbado, Garc{\'\i}a, Gil-L{\'o}pez, Molina, Benjamins
  et~al.}]{arrieta2020explainable}
\bibinfo{author}{A.~B. Arrieta}, \bibinfo{author}{N.~D{\'\i}az-Rodr{\'\i}guez},
  \bibinfo{author}{J.~Del~Ser}, \bibinfo{author}{A.~Bennetot},
  \bibinfo{author}{S.~Tabik}, \bibinfo{author}{A.~Barbado},
  \bibinfo{author}{S.~Garc{\'\i}a}, \bibinfo{author}{S.~Gil-L{\'o}pez},
  \bibinfo{author}{D.~Molina}, \bibinfo{author}{R.~Benjamins}, et~al.,
\newblock \bibinfo{title}{{Explainable Artificial Intelligence (XAI): Concepts,
  taxonomies, opportunities and challenges toward responsible AI}},
\newblock \bibinfo{journal}{Information fusion} \bibinfo{volume}{58}
  (\bibinfo{year}{2020}) \bibinfo{pages}{82--115}.
%Type = Inproceedings
\bibitem[{Xue et~al.(2021)Xue, Constant, Roberts, Kale, Al-Rfou, Siddhant,
  Barua, and Raffel}]{xue-etal-2021-mt5}
\bibinfo{author}{L.~Xue}, \bibinfo{author}{N.~Constant},
  \bibinfo{author}{A.~Roberts}, \bibinfo{author}{M.~Kale},
  \bibinfo{author}{R.~Al-Rfou}, \bibinfo{author}{A.~Siddhant},
  \bibinfo{author}{A.~Barua}, \bibinfo{author}{C.~Raffel},
\newblock \bibinfo{title}{m{T}5: A massively multilingual pre-trained
  text-to-text transformer},
\newblock in: \bibinfo{booktitle}{NAACL}, \bibinfo{year}{2021}.
%Type = Inproceedings
\bibitem[{Zhang et~al.(2020)Zhang, Kishore, Wu, Weinberger, and
  Artzi}]{zhangbertscore}
\bibinfo{author}{T.~Zhang}, \bibinfo{author}{V.~Kishore},
  \bibinfo{author}{F.~Wu}, \bibinfo{author}{K.~Q. Weinberger},
  \bibinfo{author}{Y.~Artzi},
\newblock \bibinfo{title}{{BERTScore: Evaluating Text Generation with BERT}},
\newblock in: \bibinfo{booktitle}{International Conference on Learning
  Representations (ICLR)}, \bibinfo{year}{2020}.
%Type = Article
\bibitem[{Miller(2019)}]{miller2019explanation}
\bibinfo{author}{T.~Miller},
\newblock \bibinfo{title}{{Explanation in artificial intelligence: Insights
  from the social sciences}},
\newblock \bibinfo{journal}{Artificial intelligence} \bibinfo{volume}{267}
  (\bibinfo{year}{2019}) \bibinfo{pages}{1--38}.
%Type = Book
\bibitem[{Johnson(2012)}]{johnson2012manifest}
\bibinfo{author}{R.~H. Johnson}, \bibinfo{title}{Manifest rationality: A
  pragmatic theory of argument}, \bibinfo{publisher}{Routledge},
  \bibinfo{year}{2012}.
%Type = Inproceedings
\bibitem[{Singhal et~al.(2022)Singhal, Azizi, Tu, Mahdavi, Wei, Chung, Scales,
  Tanwani, Cole-Lewis, Pfohl, Payne, Seneviratne, Gamble, Kelly, Scharli,
  Chowdhery, Mansfield, y~Arcas, Webster, Corrado, Matias, Chou, Gottweis,
  Tomasev, Liu, Rajkomar, Barral, Semturs, Karthikesalingam, and
  Natarajan}]{singhal2022large}
\bibinfo{author}{K.~Singhal}, \bibinfo{author}{S.~Azizi},
  \bibinfo{author}{T.~Tu}, \bibinfo{author}{S.~S. Mahdavi},
  \bibinfo{author}{J.~Wei}, \bibinfo{author}{H.~W. Chung},
  \bibinfo{author}{N.~Scales}, \bibinfo{author}{A.~Tanwani},
  \bibinfo{author}{H.~Cole-Lewis}, \bibinfo{author}{S.~Pfohl},
  \bibinfo{author}{P.~Payne}, \bibinfo{author}{M.~Seneviratne},
  \bibinfo{author}{P.~Gamble}, \bibinfo{author}{C.~Kelly},
  \bibinfo{author}{N.~Scharli}, \bibinfo{author}{A.~Chowdhery},
  \bibinfo{author}{P.~Mansfield}, \bibinfo{author}{B.~A. y~Arcas},
  \bibinfo{author}{D.~Webster}, \bibinfo{author}{G.~S. Corrado},
  \bibinfo{author}{Y.~Matias}, \bibinfo{author}{K.~Chou},
  \bibinfo{author}{J.~Gottweis}, \bibinfo{author}{N.~Tomasev},
  \bibinfo{author}{Y.~Liu}, \bibinfo{author}{A.~Rajkomar},
  \bibinfo{author}{J.~Barral}, \bibinfo{author}{C.~Semturs},
  \bibinfo{author}{A.~Karthikesalingam}, \bibinfo{author}{V.~Natarajan},
\newblock \bibinfo{title}{Large language models encode clinical knowledge},
\newblock in: \bibinfo{booktitle}{arXiv 2212.13138}, \bibinfo{year}{2022}.
%Type = Article
\bibitem[{Mayer et~al.(2021)Mayer, Marro, Cabrio, and
  Villata}]{mayer2021enhancing}
\bibinfo{author}{T.~Mayer}, \bibinfo{author}{S.~Marro},
  \bibinfo{author}{E.~Cabrio}, \bibinfo{author}{S.~Villata},
\newblock \bibinfo{title}{{Enhancing Evidence-Based Medicine with Natural
  Language Argumentative Analysis of Clinical Trials}},
\newblock \bibinfo{journal}{Artificial Intelligence in Medicine}
  (\bibinfo{year}{2021}) \bibinfo{pages}{102098}.
%Type = Inproceedings
\bibitem[{Garc{\'\i}a-Ferrero et~al.(2022)Garc{\'\i}a-Ferrero, Agerri, and
  Rigau}]{garcia-ferrero-etal-2022-model}
\bibinfo{author}{I.~Garc{\'\i}a-Ferrero}, \bibinfo{author}{R.~Agerri},
  \bibinfo{author}{G.~Rigau},
\newblock \bibinfo{title}{Model and data transfer for cross-lingual sequence
  labelling in zero-resource settings},
\newblock in: \bibinfo{booktitle}{Findings of the Association for Computational
  Linguistics: EMNLP 2022}, \bibinfo{year}{2022}.
%Type = Inproceedings
\bibitem[{Yeginbergenova and Agerri(2023)}]{yeginbergenova2023crosslingual}
\bibinfo{author}{A.~Yeginbergenova}, \bibinfo{author}{R.~Agerri},
\newblock \bibinfo{title}{Cross-lingual argument mining in the medical domain},
\newblock in: \bibinfo{booktitle}{arXiv 2301.10527}, \bibinfo{year}{2023}.
%Type = Inproceedings
\bibitem[{Devlin et~al.(2019)Devlin, Chang, Lee, and Toutanova}]{Devlin19}
\bibinfo{author}{J.~Devlin}, \bibinfo{author}{M.~Chang},
  \bibinfo{author}{K.~Lee}, \bibinfo{author}{K.~Toutanova},
\newblock \bibinfo{title}{{BERT:} pre-training of deep bidirectional
  transformers for language understanding},
\newblock in: \bibinfo{booktitle}{{NAACL-HLT}}, \bibinfo{year}{2019}, pp.
  \bibinfo{pages}{4171--4186}.

\end{thebibliography}

%\appendix


\end{document}